\documentclass{article}

\usepackage[nonatbib, final]{lib}
\usepackage[utf8]{inputenc} % allow utf-8 input
\usepackage[T1]{fontenc}    % use 8-bit T1 fonts
\usepackage{url}            % simple URL typesetting
\usepackage{booktabs}       % professional-quality tables
\usepackage{amsfonts}       % blackboard math symbols
\usepackage{nicefrac}       % compact symbols for 1/2, etc.
\usepackage{microtype}      % microtypography
\usepackage{caption}        % caption
\usepackage{subcaption}     % subcaption
\usepackage{graphicx}       % graphicx
\usepackage{adjustbox}      % adjustbox
\usepackage{amssymb}
\usepackage{algorithm}
\usepackage[noend]{algpseudocode}
\usepackage[pagebackref=true,breaklinks=true,colorlinks,bookmarks=false]{hyperref}
\usepackage{float}
\usepackage{enumitem}
\usepackage{amsmath}
\title{Image Labeling with Markov Random Fields and Conditional Random Fields}

\author{
Shangxuan Wu\\
Robotics Institute\\
Carnegie Mellon University\\
\texttt{ shangxuw@andrew.cmu.edu}
\And
Xinshuo Weng\\
Robotics Institute\\
Carnegie Mellon University\\
\texttt{ xinshuow@andrew.cmu.edu}
}

\makeatletter
\def\BState{\State\hskip-\ALG@thistlm}
\makeatother

\begin{document}

\maketitle
\begin{abstract}
Most existing methods for object segmentation in computer vision are formulated as a labeling task. This, in general, could be transferred to a pixel-wise label assignment task, which is quite similar to the structure of hidden Markov random field. In terms of Markov random field, each pixel can be regarded as a state and has a transition probability to its neighbor pixel, the label behind each pixel is a latent variable and has an emission probability from its corresponding state. In this paper, we reviewed several modern image labeling methods based on Markov random field and conditional random Field. And we compare the result of these methods with some classical image labeling methods. The experiment demonstrates that the introduction of Markov random field and conditional random field make a big difference in the segmentation result.
\end{abstract}

\section{Introduction}
In the field of computer vision, segmentation \cite{he2004multiscale, chen2016deeplab, chen2015rnn, koltun2011efficient}, i.e. image labeling often plays an intermediate but important role for some higher level image understanding and recognition. To some extent, the precision and accuracy of image labeling could be a bottleneck of these future work. So the challenge of achieving more find-grained segmentation result always attracts lots of research interest in the field. 

In terms of image labeling task, each individual pixel of a given image needs to be assigned one specific label from a predefined discrete label set. But it's not possible to figure out what the best label is for each pixel by just looking at its pixel value. So the context information is needed for the labeling task. Some low level information and pattern in the image is also quite useful. For example, the color and texture might give a good sense of what the object class is, because pixel with the same color and texture might has higher possibility to be assigned a same label. But the challenge is some different classes (sky and water) share same color or texture and some objects involve multiple colors (human clothing). For disambiguating these confusion, more attempts in the field has incorporated additional information such as localization property, which means pixels with the same color and also close to each other should be assigned a same label. Also many graphical models have been used for modeling this task.

In this paper, we mainly focus on modeling the image labeling task by exploring the Markov property. One could always formulate all pixels in the image as an undirected graphical model. More specifically, to limit the scope of this paper, we only consider graphical model with 2D Markov property. That means the state of each pixel can only be determined by the state from a pre-defined set of pixel, such as four immediately connected neighbor pixels. In terms of labeling task, each pixel needs to be associated with one specific label from label classes. This refers to emission probability from the perspective of hidden Markov random field. Because of this similarity between Markov random field and image labeling, we could simply transfer the image labeling task to model the full joint probability of the image and corresponding labels. In addition to the basic review of MRFs method, we focus more on conditional random field, which only models the conditional probability of labels given the image. This decreases a huge mount of parameter to be estimated in the model. Our comparative experimental results demonstrate that CRFs allow us to get more robust segmentation result with easier and faster inference process. Compared to the classical methods, the experiments validate our claim that model imposing the Markov property is superior to the traditional methods using only low level image information.

\section{Related Works}
\label{sec:related}
One response to image labeling task is traditional unsupervised methods involving all kinds of low level information. By measuring color \cite{Coombes2016color} and texture similarity, lots of attempts involving clustering methods such as k-means \cite{Tatiraju2008kmeans} and mean-shift \cite{dorin2002meanshift} try to group the local region with higher similarity and assign them with a same label. For computational convenience, superpixel \cite{Li2012superpixel, Wang2009superpixel} approach is often used as pre-processing step to group the potential similar pixel together. Then all previous approach works on the superpixel group instead of individual pixel. But the problem of these approach occurs due to the use of only local information. So research work starts to combine global and local context to get a better sense of semantic information in the image. In \cite{Sumengen2005} a multi-scale structure is used for segmenting the image based on edge detection with different scale level. In \cite{Liu2015parsenet, Rabinovich2007context}, combination of global context and local informationis proved to be very useful in segmentation task.

Another set of research work for solving image labeling task in vision is to couple one or more other high-level and related tasks with segmentation. In other words, additional high-level information is used for solving labeling task simultaneously. For example, in \cite{Liu2016detection, Dong2014detection, Gupta2014detection, Hariharan2014detection, Hariharan2015localization} multi-task learning has been explored to solve object detection and segmentation simultaneously. In \cite{Chen2016attention} attention model is used to improve the performance of segmentation. Other related tasks such as edge detection, depth estimation and surface normal are also leveraged to couple with segmentation and achieve better result in \cite{Eigen2014depth, Chen2015edge}. Even human interaction \cite{boykov2001interactive} with the input is very useful to provide additional information for better segmentation result.

The first time that Markov random field is introduced for segmentation is in \cite{zhang2001segmentation}, where it's only for the application of medical image. Then the attempt of modeling segmentation task with Markov random field is largely growing due to the similarity of their representation. But learning and inference for Markov random field is quite computationally expensive so that lots of approximation and learning scheme is invented for obtaining the robust segmentation result. In addition to Markov random field, one point worth to highlight is conditional random field, which in \cite{he2004multiscale, Sutton2004crf, Lafferty2001crf} is heavily used for segmentation in the recent decades. The benefit of conditional random field is that, for image labeling task, we actually don't care about the joint distribution of image pixel. So modeling only the label conditioned on image observation is enough and will decrease huge mount of parameters to estimate. Most recently, with the revolution of deep learning, conditional random field is usually used as a post-precessing step in \cite{chen2016deeplab} to refine the segmentation result produced by deep convolutional neural network. More attempt \cite{chen2015rnn} etries to incorporate conditional random field into network architecture instead of post processing and achieve better result.

\section{Random Field for Image Labeling}
In this section, we review basics about how to model the segmentation task using Markov random field and conditional random field respectively. Obviously, image labeling needs information across many neighboring pixels, which we called contextual or local information. We can reasonable assume that the nearby pixels are more likely to be assigned a same label. With the use of Markov random fields, we typically formulated a probabilistic generative framework modeling the joint probability of the image and its corresponding labels. This allows our model to locally smooth the assigned labels. But the underlying generative model by MRFs is actually more complicated than what we need, because we are only interested in the posterior labels given the observed image instead of full joint distribution over all pixel values and its corresponding labels. Parameter estimation for the full joint generative model based on MRFs is very hard due to huge mount of parameters. So modeling only the conditional probability of the labels, namely conditional random fields, will significantly boost the speed. Actually, CRF models labels as random variables that forms Markov random field when conditioned upon a global observation. So MRF is an generative model but CRF is a discriminative model.

\subsection{Modeling}

Concretely, in the fully connected pairwise CRF model, we will define the energy function for label assignment of each pixel here as follows,
\begin{equation}
\label{eq:solve}
E(x)=\sum_i\psi_u(x_i)+\sum_{i<j}\psi (x_i,x_j)
\end{equation}
where $\psi_u$ is the unary energy component and $\psi_p$ represents the neighborhood pairwise energy component,
\begin{equation}
\label{eq:binary}
\psi_p(x_i,x_j)=\mu(x_i,x_j)\sum_{m=1}^Mw^{(m)}k_G^{(m)}(f_i,f_j)
\end{equation}
Minimizing the CRF energy $E(x)$ above yields the most probable label assignment $x$ for the given image.

\subsection{Inference}
In this section, we introduce several algorithms for solving the inference problem given a conditional random field model with known parameters. These are all iterative models, and one can either use unary potential or use other prior knowledge to initialize the inference.

\subsubsection{Iterated Conditional Modes (ICM)}
This is simply a greedy algorithm, but when unary clique potential dominates the potential, ICM could reach global optimum. It optimizes the potential by iteratively maximizing the conditional probability of $P(one variable|other variables)$.

\subsubsection{Simulated Annealing}
Simulated Annealing is a classical technique for optimization. It simulates the process of annealing to find global minimals or maximals from local minimals or maximals.
\begin{algorithm}[H]
\caption{Simulated Annealing algorithm}
% \label{euclid}
\begin{algorithmic}[1]
\Procedure{Initialization}{}
\State Initialize the state $s$ to $s_0$
\EndProcedure
\Procedure{Loop for $n = N$ times}{}
\State $T \gets temparature(\frac{T}{T_{max}})$
\State Get a new state $s_{new}$ from temperature $T$
\State If $f(s_{new})>f(s_{old})$, $s \gets s_{new}$
\EndProcedure
\Procedure{Output}{}
\State Output the state $s_{n=N}$
\EndProcedure
\end{algorithmic}
\end{algorithm}
The random process of getting temperature could help avoiding stuck of the local minimal or maximal, and start finding the new and global extrema. Also, the temperature would be lower with the time increasing, and the state would be more and more stable.

\subsubsection{Belief Propagation}
Belief propagation, also known as "sum-product message passing", provides exact solution when there are no loop in the graph. Otherwise, belief propagation provides approximate (but often good) solution \cite{james2009slide}.

In BP, the estimated marginal probabilities are called beliefs. It updates message until convergence, and then calculates the beliefs. Here, we illustrate the message update of pairwise MRF in Equation \ref{eq:solve} as the following schematic diagram:
\begin{figure}[H]
  \centering
\begin{subfigure}{0.7\textwidth}
  \includegraphics[width=0.9\linewidth]{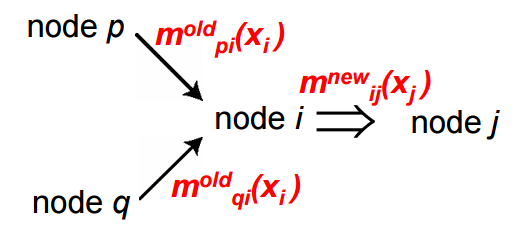}
%   \label{fig:sfig1}
\end{subfigure}%
\caption{Illustration of message passing step in BP.}
% \label{fig:fig}
\end{figure}

where $m_{ij}(x_j)$ represents message from node $i$ to node $j$. And our belief could be represented as
\begin{equation}
b_i(x_i) \propto g_i(x_i)\prod_{k \in Nbd(i)} m_{ki}x(i)
\end{equation}
when the message converges.
\subsubsection{Graph Cuts}
\begin{figure}[H]
\minipage{0.5\textwidth}
  \includegraphics[width=\linewidth]{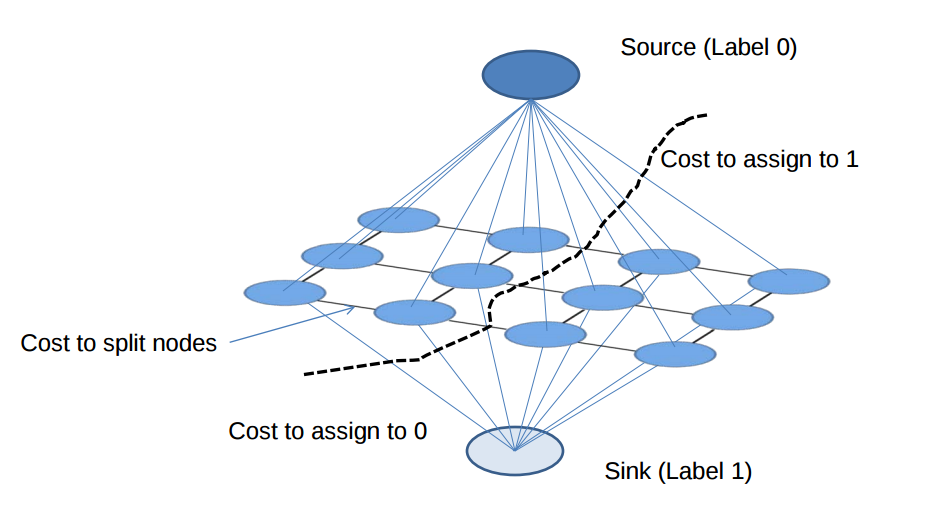}
  \caption*{Graph Cut Initial State.}
  \label{fig:awesome_image1}
\endminipage\hfill
\minipage{0.5\textwidth}
  \includegraphics[width=\linewidth]{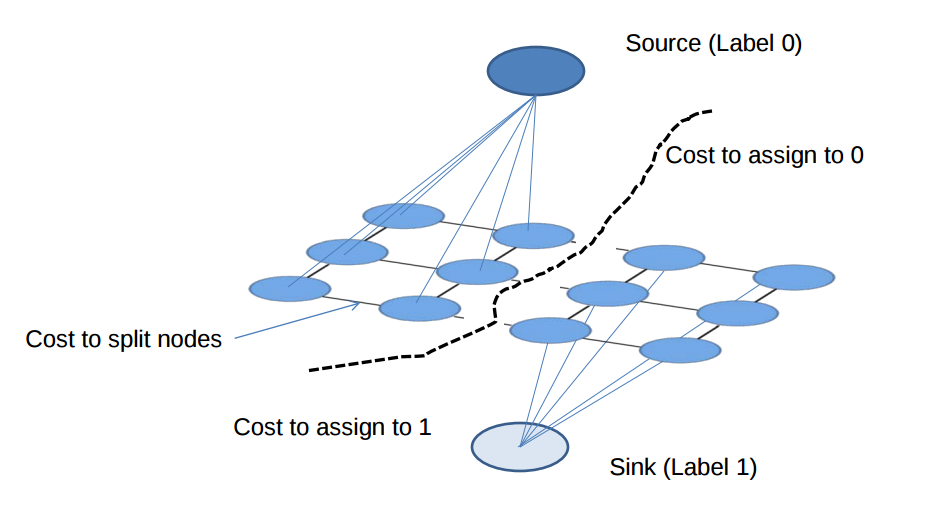}
  \caption*{Graph Cut Ending State.}
%   \label{fig:awesome_image2}
\endminipage\hfill
\end{figure}
\cite{greig1989exact} shows that, if the pairwise potentials of a \textbf{two-label} pairwise MRF could be defined as an Ising model, then we could get an accurate MAP solution by solving a mincut problem. By this theorem, the total energy $C(x)$ could be written as
\begin{equation}
C(x)=\sum_{i=1}^nx_imax(0,1-\lambda_i)+\sum_{i=1}^n(1-x_i)max(0,\lambda_i)+\frac{1}{2}\sum_{i=1}^n\sum_{j=1}^n\beta_{ij}(x_i-x_j)^2
\end{equation}
where $x=(x_1,...,x_n)$ means a binary image and capacity $c_{si} = \lambda_i$.

\subsubsection{Mean-Field Algorithm}
Mean-field algorithm is used to approximate the maximum a posterior marginal inference. It uses Gibbs distribution to initialize, and performs loop to minimize the CRF total energy $Q$. Message passing step computes the  weighted Gaussian, and the following two steps converts it to the binary term of equation \ref{eq:binary}. The algorithm is summarized in \ref{al:mean}. In \cite{chen2015rnn}, mean-field algorithm is proved to be able to be fitted as layers of recurrent neural network.
\begin{algorithm}[H]
\caption{Mean-field inference algorithm \cite{chen2015rnn}}
% \label{euclid}
\begin{algorithmic}[1]
\Procedure{Initialization}{}
\State $Q_i(l) \gets \frac{1}{Z_i}exp(U_i(l))$ for all $i$
\EndProcedure
\Procedure{Loop Until Converge}{}
\State $Q_i^{(m)}(l) \gets \sum_{j\neq i}k^{(m)}(f_i,f_j)Q_j^{(l)}$ for all $m$ (Message Passing)
\State $Q_i^{(l)} \gets \sum_mw^{(m)}Q_i(l)$(Weighting Filter Outputs)
\State $Q_i^{(l)} \gets \sum_{l'\in \textbf{L}}\mu(l,l')Q_i(l')$ (Compatibility Transform)
\State $Q_i^{(l)} \gets U_i(l)-Q_i(l)$ (Adding Unary Potentials)
\State $Q_i^{(m)} \gets \frac{1}{Z_i}exp(Q_i(l))$ for all $m$ (Normalizing)
\EndProcedure
\end{algorithmic}
\label{al:mean}
\end{algorithm}

\subsection{Parameter Estimation}
Given the model, we need to find out the appropriate parameters for future inference. But the learning process of CRFs is very complicated. Given a $N \times N$ pixel picture, we can get a $N^2$-D space. How do we model the internal parameters?  Data-driven optimization algorithms are entailed in this learning process. Here we won't cover much of the learning process. Please see \cite{wang2000mrf} for reference of one learning algorithm called Markov Chain Monte Carlo (MCMC) method.

\section{Experimental Results}
\subsection{Introduction of MRF/CRF-based Methods}
In section \ref{sec:related} we introduced many image labeling methods. Here we make a detailed introduction to three of them: Hidden MRF and its Expectation-Maximization Algorithm \cite{zhang2001segmentation}, CRF-as-RNN \cite{chen2015rnn} and DeepLab \cite{chen2016deeplab}.
\subsubsection{Hidden MRF and its Expectation-Maximization Algorithm}

In \cite{zhang2001segmentation}. A method for the segmentation of magnetic resonance images is proposed. The main algorithm for this paper is shown below:

According to MAP criterion, we seek the labeling \textbf{x*} that satisfy the following rule:
\begin{equation}
    x^* = arg\max_x \{P(y|x,\Theta)P(x)\}
\end{equation}
Therefore, \cite{zhang2001segmentation} uses EM algorithm to estimate the parameter set $\Theta = \{\theta|l\in L\}$.
\begin{algorithm}[H]
\caption{Hidden MRF-EM algorithm}
% \label{euclid}
\begin{algorithmic}[1]
\Procedure{Start}{}
\State Initialize parameter set $\Theta \gets \Theta^{(0)}$
\EndProcedure
\Procedure{Loop}{}
\State \textbf{E step}: Calculate the conditional expectation $Q(\Theta|\Theta^t)$.
\State \textbf{M step}: Maximize $Q(\Theta|\Theta^t)$ and get $\Theta^{t+1}$.
\EndProcedure
\end{algorithmic}
\end{algorithm}
Then an iterative algorithm is used to minimize the total posterior energy:
\begin{equation}
    x^* = arg\min_{x\in X}\{U(y|x,\Theta)+U(x)\}
\end{equation}
\begin{figure}[H]
\begin{subfigure}{0.33\textwidth}
  \centering
  \includegraphics[width=0.8\linewidth]{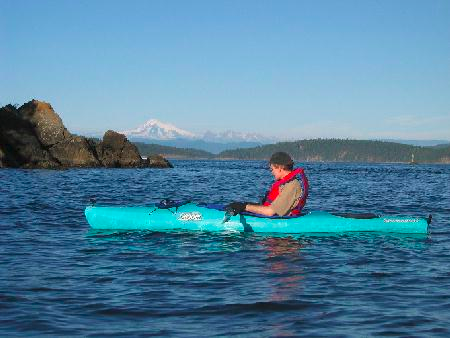}
  \caption*{Original image.}
%   \label{fig:sfig1}
\end{subfigure}%
\begin{subfigure}{0.33\textwidth}
  \centering
  \includegraphics[width=0.8\linewidth]{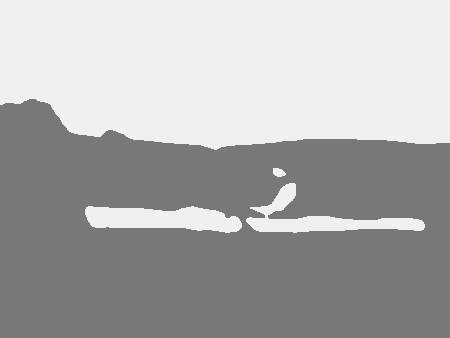}
  \caption*{HMRF-EM algorithm result.}
%   \label{fig:sfig1}
\end{subfigure}%
\begin{subfigure}{0.33\textwidth}
  \centering
  \includegraphics[width=1\linewidth]{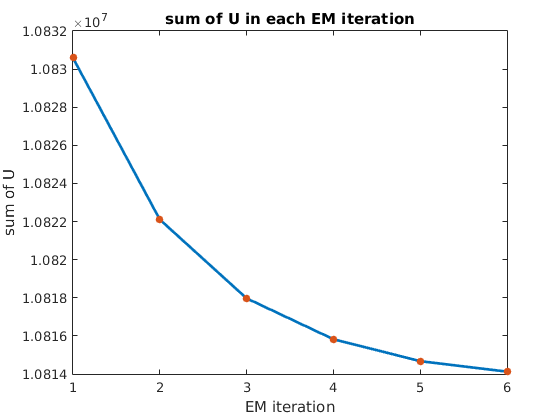}
  \caption*{Sum of energy in each EM iteration.}
%   \label{fig:sfig1}
\end{subfigure}%
\caption{HMRF-EM algorithm results.}
% \label{fig:fig}
\end{figure}

\subsubsection{CRF as RNN}

\begin{figure}[H]
%\minipage{1\textwidth}\centering
%  \includegraphics[width=1\linewidth]{image/CRFasCNN.png}
%  \caption{CRF facilitated as a CNN neuron.\cite{chen2015rnn}}\label{fig:awesome_image2}
%\endminipage\vfill
\minipage{1\textwidth}%
	\centering
  \includegraphics[width=0.55\linewidth]{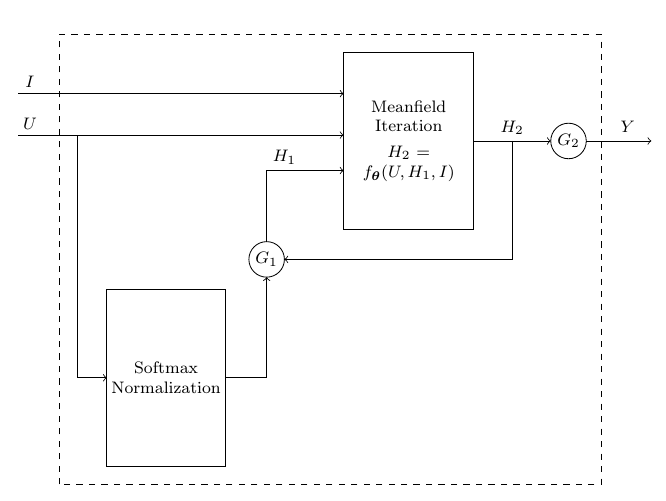}
  \caption{CRF facilitated as a RNN neuron.\cite{chen2015rnn}}
%   \label{fig:awesome_image3}
\endminipage
\end{figure}

In \cite{koltun2011efficient}, the message passing step is computed by a stack of convolutions. Thus in \cite{chen2015rnn}, these computation is then transferred to a stack of layers in the recurrent neural network by decomposing equation \ref{eq:solve}. These neurons are designed to perform the following steps:
\begin{itemize}
    \item Initialization
    \item loop
    \begin{itemize}
      \item Message Passing
    \item Weighting Filter Outputs
    \item Compatibility Transform
    \item Adding Unary Potentials
    \item Normalizing 
    \end{itemize}
\end{itemize}
These steps composes to the following equation:
\begin{equation}
U(X, I) = \sum_i \psi_u(x_i, I_i) + \sum_{i<j} \mu_p(x_i,x_j)\sum_{m=1}^Mw^{(m)}k_G^{(m)}(f_i, f_j)
\end{equation}

Denote the unary energy as $U$, parameters $\{w^{(m)},\mu(l,l')\}$ as $\textbf\{\theta\}$, 
then the RNN cell follows these equations could be denoted as
\begin{equation}
H_1(t)=\begin{cases}
               softmax(U), t=0\\
               H_2(t-1), 0<t\leq T\\
            \end{cases}
\end{equation}

\begin{equation}
    f_\theta (U,H_1(t),I), 0\leq t<T,
\end{equation}

\begin{equation}
    Y(t)=\begin{cases}
               0, 0\leq t<T\\
               H_2(t), t=T \\
            \end{cases}
\end{equation}
\subsubsection{DeepLab}
Proposed in 2015, DeepLab \cite{chen2016deeplab} should be the most famous CNN-based image labeling algorithm that incorporates conditional random fields. As a complementary component of CNN labeling results, this algorithm chose fully connected CRFs to enhance the segmentation result. The basic work flow of this algorithm is as following. We are not going to discuss the implementation details of CNN, but only focusing on the CRFs part.
\begin{figure}[H]
\minipage{1\textwidth}\centering
  \includegraphics[width=0.8\linewidth]{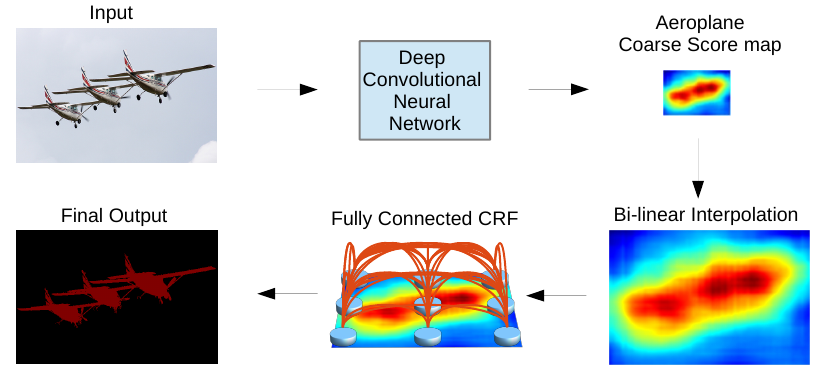}
  \caption{The basic work flow of DeepLab algorithm.\cite{chen2016deeplab}}
  \label{fig:awesome_image2}
\endminipage

\minipage{1\textwidth}\centering
  \includegraphics[width=0.8\linewidth]{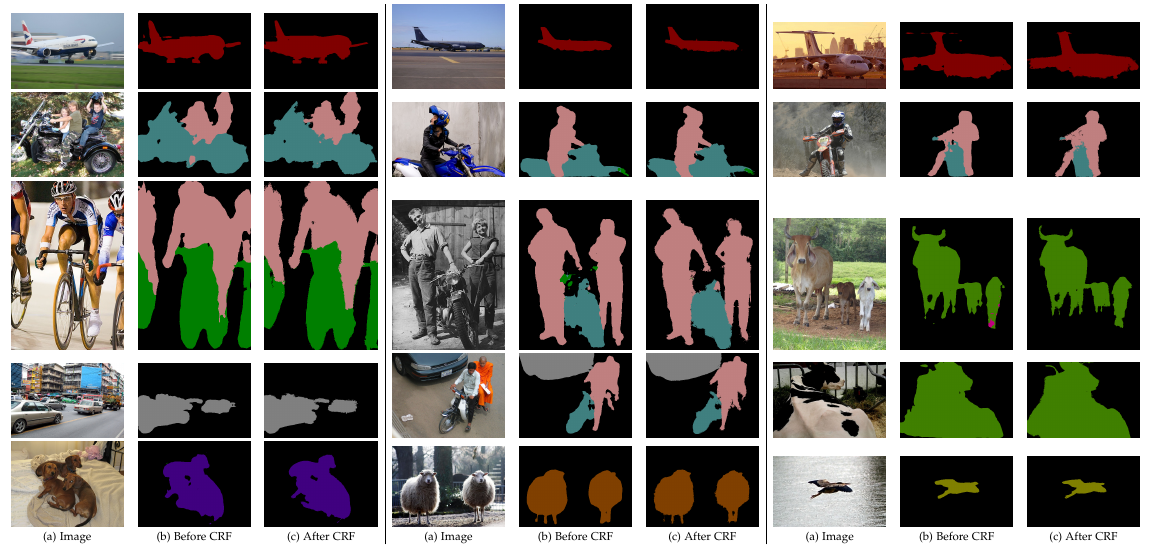}
  \caption{CNN outputs before and after CRF.}
\endminipage
\end{figure}

% Please regard to section \ref{sec:DenseCRF} for detail.

\subsection{Evaluations}
In this section, we evaluate the different usage of Markov random field and conditional random field by several experiments.

\subsubsection{MRF as Edge Detection}
Comparing to image labeling, edge detection is an easier task. In the following experiment, we show the importance of MRFs in edge detection.
\begin{figure}[H]
\begin{subfigure}{1\textwidth}
  \centering
  \includegraphics[width=.8\linewidth]{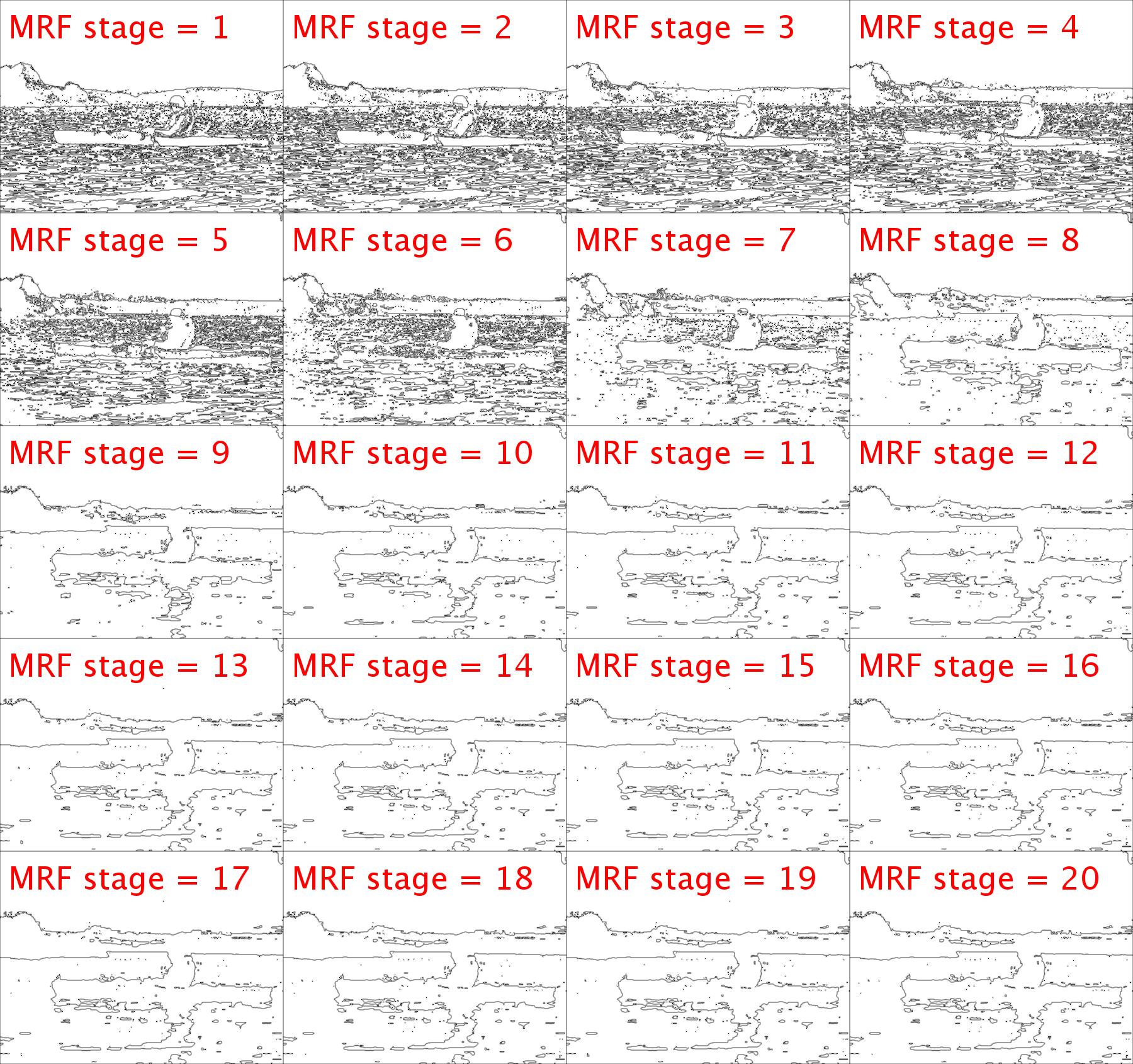}
%   \label{fig:sfig1}
\end{subfigure}%
\caption{MRF iterations could improve the accuracy of edge detection.}
% \label{fig:fig}
\end{figure}
This is an illustration of how MRF could improve the accuracy of edge detection by iterations. In the first row, we could see that the waves on the lake are separated into different small pieces, which is not what we wan. After 
10+ iterations, noise are reduced and "semantic" edges are preserved. It is easy to see that MRF could make the edge a more semantically coherent.

\subsubsection{DenseCRF for improving unsupervised segmentation}
\begin{figure}[H]
\minipage{0.24\textwidth}
  \includegraphics[width=1\linewidth]{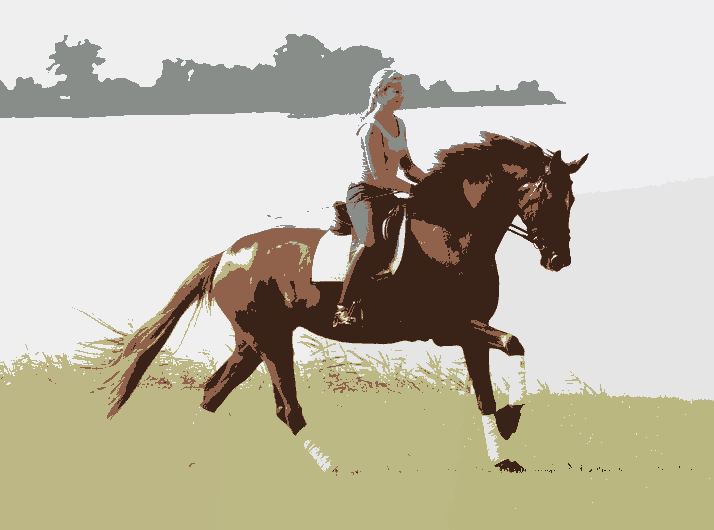}
  \caption*{K-means initialization.}
%   \label{fig:awesome_image2_for_k}
\endminipage\hfill
\minipage{0.24\textwidth}%
  \includegraphics[width=1\linewidth]{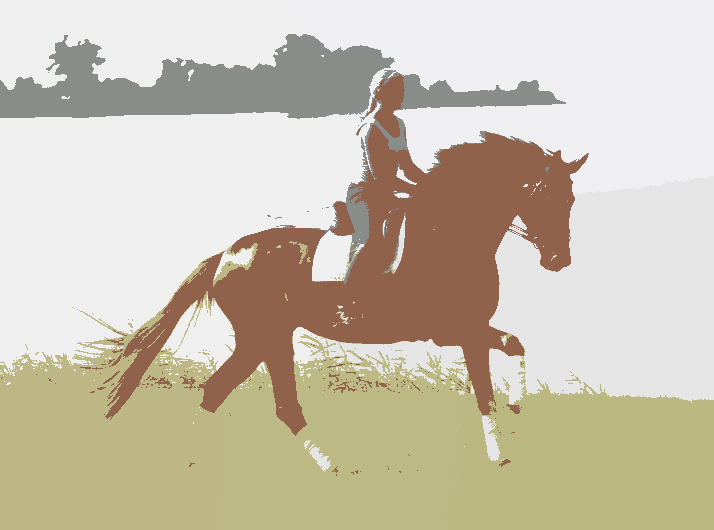}
  \caption*{5 iterations.}
%   \label{fig:awesome_image3}
\endminipage
\hfill
\minipage{0.24\textwidth}
  \includegraphics[width=1\linewidth]{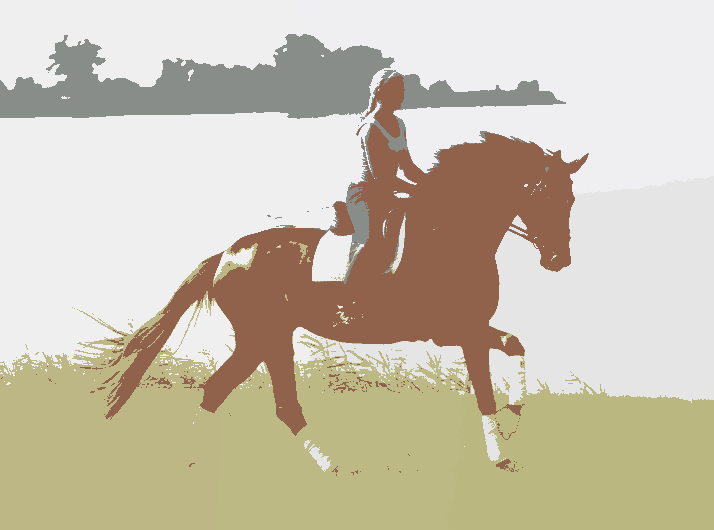}
  \caption*{10 iterations.}
%   \label{fig:awesome_image2}
\endminipage\hfill
\minipage{0.24\textwidth}%
  \includegraphics[width=1\linewidth]{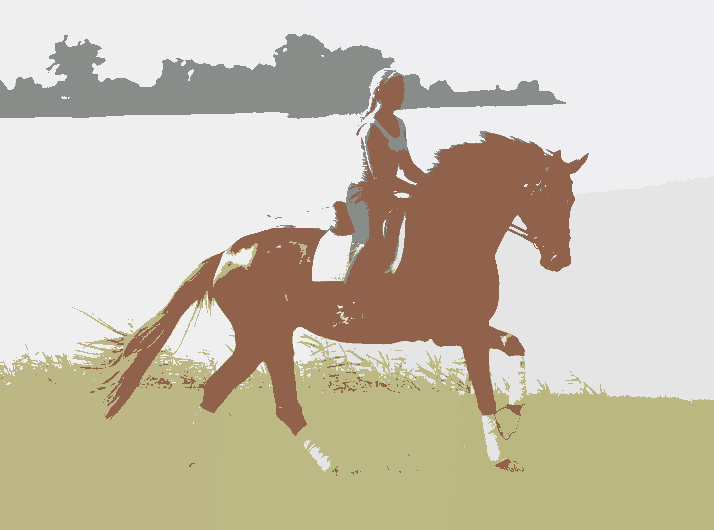}
  \caption*{100 iterations.}
%   \label{fig:awesome_image3}
\endminipage
\caption{Result of K-Means segmentation and result after CRFs post processing.}
\label{fig:iterations}
\end{figure}

As we all know, K-Means is one of the classical (and best-known) image labeling solutions. However, when given complex images, k-means clustering could not reach an accurate synthetic segmentation result. We compare the segmentations results from k-means with the result after CRFs post processing. Result is summarized in Figure \ref{fig:iterations}.

In the first figure, original K-Means segmentation ($K=5$) result is shown. Different color means different categories. You can see that the horse is seperated into different classes, which is not acceptable. But after several iterations of CRF post-processing, the several classes on horse is combined together and formed a whole segmentation.

Also notice that, the front foot of horse is discarded by K-Means, but reconstructed by CRF method. (Best viewed in PDF)

\subsubsection{DenseCRF for improving supervised segmentation algorithms}

We use \cite{chen2015rnn} to show that CRF method could significantly improve the result of supervised learning.
We get the medium result of CRF-as-RNN network and pass it to a CRF algorithm. Clearly, DenseCRF could tighten the image boundary and shrink the wrong segmentation to a proper position. (Notice how hair segmentation was shrinking!)
\begin{figure}[H]
\minipage{0.3\textwidth}
  \includegraphics[width=\linewidth]{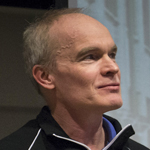}
  \caption*{Original image.}
%   \label{fig:awesome_image1}
\endminipage\hfill
\minipage{0.3\textwidth}
  \includegraphics[width=\linewidth]{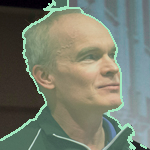}
  \caption*{Medium result of \cite{chen2015rnn} before RNN layer.}
%   \label{fig:awesome_image2}
\endminipage\hfill
\minipage{0.3\textwidth}%
  \includegraphics[width=\linewidth]{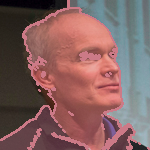}
  \caption*{With DenseCRF \cite{koltun2011efficient}.}
%   \label{fig:awesome_image3}
\endminipage
\end{figure}
\subsubsection{Comparison of MRF and CRF methods}

Comparing to MRFs, CRFs models the conditional distribution instead of joint distribution, since it is very difficult and unnecessary to get the full joint distribution over P(Labels, Image). CRFs reduce the time complexity from $O(n^k)$ to $O(n)$. 

\begin{figure}[H]
\begin{subfigure}{.24\textwidth}
  \centering
  \includegraphics[width=.95\linewidth,height=1in]{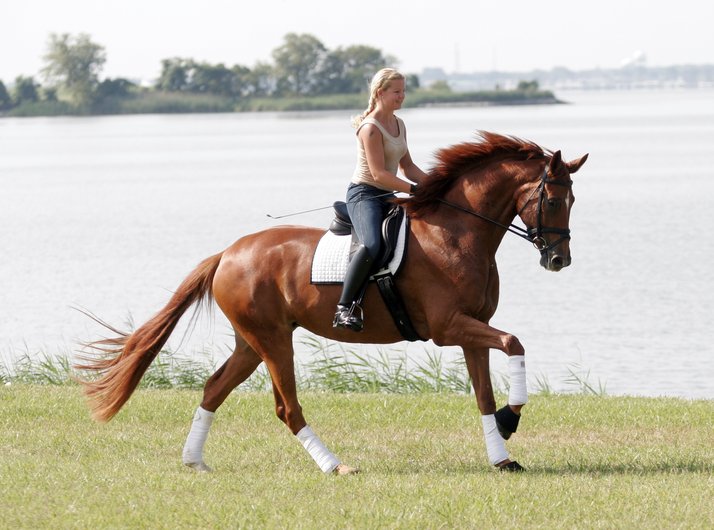}
  \caption*{{Original image}}
\end{subfigure}%
\begin{subfigure}{.24\textwidth}
  \centering
  \includegraphics[width=.95\linewidth,height=1in]{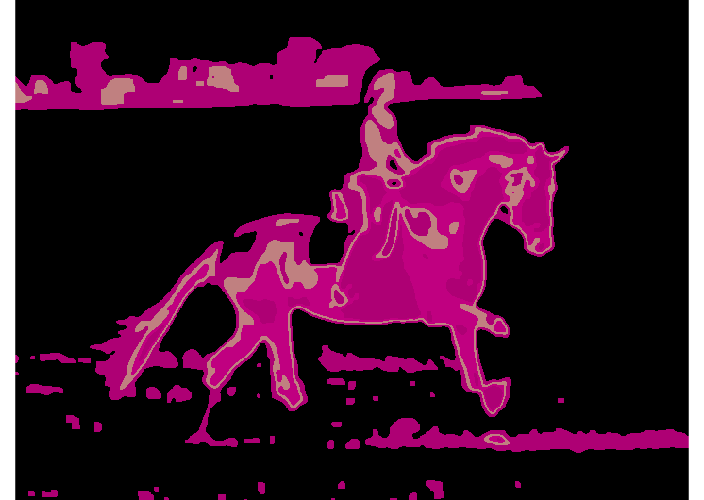}
  \caption*{{MRF}}
\end{subfigure} 
\begin{subfigure}{.24\textwidth}
  \centering
  \includegraphics[width=.95\linewidth,height=1in]{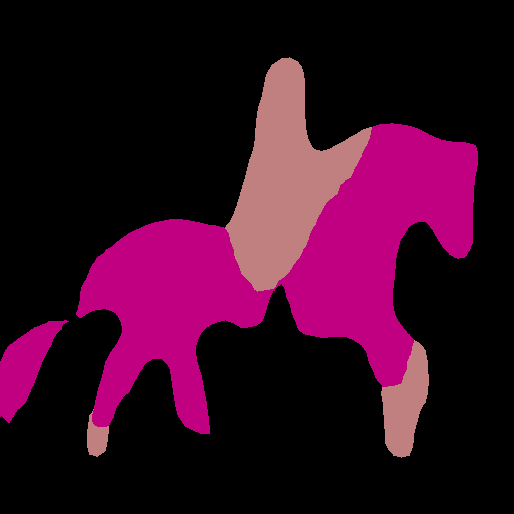}
  \caption*{{CRFasRNN \cite{chen2016deeplab}}.}
\end{subfigure}
\begin{subfigure}{.24\textwidth}
  \centering
  \includegraphics[width=.95\linewidth,height=1in]{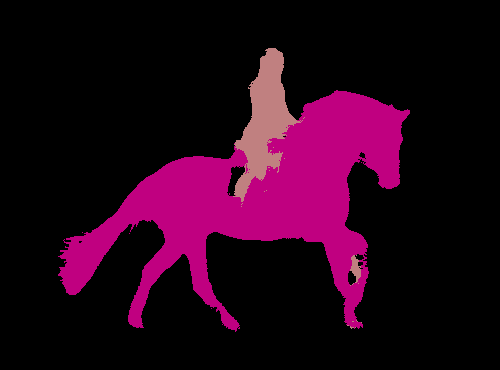}
  \caption*{{DeepLab(CRF) \cite{chen2015rnn}.}}
\end{subfigure}
\caption{Comparison between CRF and MRF methods}
\label{fig:aaa}
\end{figure}
In Microsoft COCO Image Segmentation Challenge \cite{lin2014microsoft}, both \cite{chen2015rnn} and \cite{chen2016deeplab} have achieved good rankings. Here we use these two methods as a representation of CRF image labeling methods.

In Figure \ref{fig:aaa}, we showed a comparison of how CRFs Methods outperforms MRFs in image labeling tasks. Result from CRFs are much more complete and smoother for semantic information. Also, it is far better than MRF in terms of time complexity.

\section{Conclusion}

In this paper, we review the basic concept of Markov random field and conditional random field. Also, we inspected some details in the CRFs optimization. And comparative and ablative analysis is conducted to prove the benefit of conditional random field.

In conclusion, MRFs and CRFs could significantly enhance the accuracy of image labeling in that it could make the edges smoother and more coherent to semantic objects. CRFs performs better because it models the conditional distribution instead of joint distribution. It would also reduce the time complexity for computing and make the inference more robust. 

%%----------------------------------------------------------------------------------------
%%	BIBLIOGRAPHY
%%----------------------------------------------------------------------------------------
{
\small
\bibliographystyle{ieee}
\bibliography{bib}
}

\end{document}